%% file: main.tex
\documentclass[10pt,twocolumn,letterpaper]{article}

\usepackage{cvpr}              

\input{preamble}

\definecolor{cvprblue}{rgb}{0.21,0.49,0.74}
\usepackage[pagebackref,breaklinks,colorlinks,allcolors=cvprblue]{hyperref}

\usepackage{enumitem} 
\usepackage[linesnumbered,ruled,vlined]{algorithm2e}
\usepackage{times}
\usepackage{latexsym}
\usepackage{xspace}

\usepackage{amsmath}   
\usepackage{amssymb}   
\usepackage{amsfonts}  
\usepackage{tcolorbox}  

\usepackage[T1]{fontenc}

\usepackage[utf8]{inputenc}

\usepackage{microtype}

\usepackage{inconsolata}
\usepackage{color,colortbl}
\usepackage{xcolor}
\makeatletter
\renewcommand\@makefnmark{%
  \hbox{\textsuperscript{\normalfont\color{black}\@thefnmark}}%
}
\makeatother

\usepackage{graphicx}
\usepackage{booktabs}
\usepackage{multirow}

\title{Active Video Perception: Iterative Evidence Seeking \\ for Agentic Long Video Understanding}

\author{
Ziyang Wang$^{1,2}$\thanks{Work done during internship at Salesforce.} \quad
Honglu Zhou$^{1}$ \quad
Shijie Wang$^{1}$ \quad
Junnan Li$^{1}$ \quad
Caiming Xiong$^{1}$ \\
Silvio Savarese$^{1}$ \quad
Mohit Bansal$^{2}$ \quad
Michael S. Ryoo$^{1}$ \quad
Juan Carlos Niebles$^{1}$ \\
$^{1}$Salesforce AI Research \\
$^{2}$University of North Carolina at Chapel Hill
\\
{{ \tt \normalsize \href{https://activevideoperception.github.io/}{\textcolor{blue}{https://activevideoperception.github.io/}} }}
}

\begin{document}
\maketitle
\input{sec/0_abstract}    
\input{sec/1_intro}
\input{sec/2_related_works}

\input{sec/3_method}
\input{sec/4_exp_setup}
\input{sec/5_results}

\input{sec/6_conclusion}
{
    \small
    \bibliographystyle{ieeenat_fullname}
    \bibliography{main}
}

\input{sec/X_suppl}

\end{document}

%% file: preamble.tex
\newcommand{\model}{\textsc{AVP}\xspace}

\usepackage{amsmath}

\usepackage{array,xcolor}
\newcolumntype{M}{>{\centering\arraybackslash}m{1.6cm}}

\usepackage{booktabs}
\usepackage{makecell}
\usepackage{threeparttable}

\usepackage[most]{tcolorbox}
\tcbset{
  promptstyle/.style={
    colback=black!2,
    colframe=black!30,
    boxrule=0.4pt,
    arc=1.5pt,
    left=4pt,right=4pt,top=4pt,bottom=4pt
  }
}

%% file: sec/0_abstract.tex
\begin{abstract}

Long video understanding (LVU) is challenging because answering real-world queries often depends on sparse, temporally dispersed cues buried in hours of mostly redundant and irrelevant content.
While agentic pipelines improve video reasoning capabilities, prevailing frameworks rely on a query-agnostic captioner to perceive video information, which wastes computation on irrelevant content and blurs fine-grained temporal and spatial information.
Motivated by active perception theory, we argue that LVU agents should actively decide what, when, and where to observe, and continuously assess whether the current observation is sufficient to answer the query.
We present \textbf{Active Video Perception} (\model{}), an evidence-seeking framework that treats the video as an interactive environment and acquires compact, query-relevant evidence directly from pixels.
Concretely, \model{} runs an iterative plan–observe–reflect process with MLLM agents. 
In each round, a planner proposes targeted video interactions, an observer executes them to extract time-stamped evidence, and a reflector evaluates the sufficiency of the evidence for the query, either halting with an answer or triggering further observation. 
Across five LVU benchmarks, \model{} achieves highest overall accuracy with significant improvements.
Notably, \model{} outperforms the best agentic method by \textbf{5.7\%} in average accuracy while only requires \textbf{18.4\%} inference time and \textbf{12.4\%} input tokens. 
\end{abstract}

%% file: sec/1_intro.tex
\section{Introduction}
\label{sec:intro}

\begin{figure}[t!]
  \centering
  \includegraphics[width=\columnwidth]{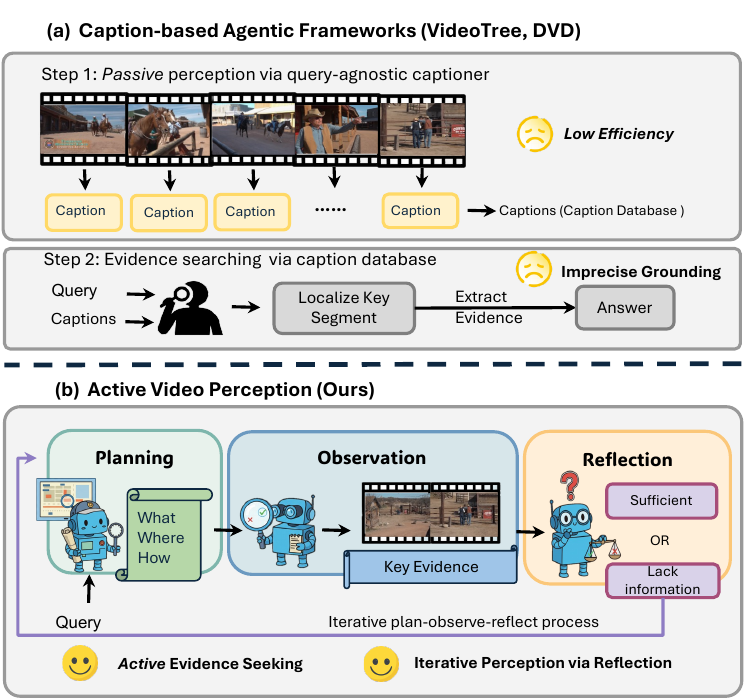}
  \caption{\textbf{Motivation of Active Video Perception}. Prior methods follow a \textit{passive} perception paradigm which leverage query-agonistic captioner to perceive the video information, leading to low efficiency and imprecise visual grounding. Instead, we \textit{actively} perceive query-relevant content by treating the long video as an interactive environment to be explored in a goal-directed manner.  }
  \label{fig:motivation}
  \vspace{-15pt}
\end{figure}

From streaming platforms to TV programs, video has become a primary medium for capturing and conveying information. 
However, long video understanding (LVU) remains challenging because it demands the ability to localize and integrate sparse, temporally dispersed cues across long time spans.
Although recent multimodal large language models (MLLMs) \citep{comanici2025gemini25pushingfrontier, openai2025gpt5, Qwen2.5-VL, wang2025internvl3_5, guo2025seed15vltechnicalreport, li2023blip2, pmlr-v162-li22n,li2025ariaopenmultimodalnative} substantially improve visual recognition, naively applying them to densely sampled, full-length videos is both computationally costly and brittle for complex queries: most video tokens are redundant, while the brief, localized evidence that actually matters is diluted or overlooked in the long sequence.

These limitations have motivated a recent surge of agentic approaches for long video understanding~\citep{zhang2023simple, wang2024videoagentlongformvideounderstanding, Yeung_2016_CVPR, wang2024videotree}.
Rather than treating the video as a single monolithic input, these methods use LLMs to orchestrate perception and reasoning over the video through planning. 
However, recent leading methods \citep{zhang2025deepvideodiscoveryagentic, zuo2025videolucydeepmemorybacktracking, shen2025vgentgraphbasedretrievalreasoningaugmentedgeneration} still rely on captioners to convert visual information into text space as the primary interface for LLM reasoning and tool calling.
This caption-based framework leverages LLMs’ strengths in text processing but introduces two inherent limitations:
\begin{enumerate}[leftmargin=1.5em]
\item \textbf{High Computational Cost:}
Query-agnostic captioning generates large amounts of irrelevant information, expending computation on unrelated content and resulting in low efficiency.
\item \textbf{Imprecise Grounding via Captions:} 
Existing approaches use captions to localize key events, which may discard fine-grained temporal and spatial cues and weaken causal tracing.
\end{enumerate}
These limitations underscore the need for an agentic framework that adaptively focuses on informative video regions, seeks query-related evidence directly over video pixels while maintaining high efficiency. 

We take inspiration from how humans inspect long videos: we do not need to watch every frame; instead, we plan our observation based on the query. 
For instance, given a question about a specific plot, we first skim the video for coarse cues (plot localization), then take targeted observation by focusing on the key video regions for detail clues.
Active perception theory \citep{bajcsy1988active,aloimonos2013active, bajcsy2016revisitingactiveperception} formalizes this behavior: \emph{``An agent is an active perceiver if it knows why it wishes to sense, and then chooses what to perceive, and determines how, when and where to achieve that perception''}. 
Even though active perception concept is mainly used in robotics domain \citep{sripada2025sceneexplorationvisionlanguagemodels, xiong2025visionactionlearningactive, shang2023activevisionreinforcementlearning}, 
We argue that agentic LVU frameworks can similarly benefit from query-driven, temporally grounded observation that decides what, when, and where to look, while continually assessing whether the accumulated evidence is sufficient for the query or whether further observation is required.

Building on this view, we propose \textbf{Active Video Perception} (\model{}), an agentic evidence seeking framework for long video understanding. 
As shown in \cref{fig:motivation}, rather than passively perceiving the video by captioning, \model{} treats the video as an interactive environment and 
actively decides what/where/how to observe the video to acquire the query-related information. 
This targeted observation design allows \model{} to focus on the key informative segments, avoid redundant processing over static or irrelevant content, ultimately improving both efficiency and reliability on complex long-horizon queries.

Since complex queries often depend on sparse or ambiguous cues that cannot be resolved in a single round of observation, \model{} adopts an iterative plan–observe–reflect process with MLLM agents.
In each round, a planner proposes targeted interactions with the video by deciding what to inspect, where to focus, and at what granularity. 
Then, a observer executes these plans to extract compact, time-stamped evidence. 
Finally, a reflector evaluates the query-sufficiency of the extracted evidence and decide whether additional round of observation is needed. 
If the extracted evidence is insufficient, it appends the current plan, evidence, and justification to the running history to guide the planner in deciding the next plan. 
This closed-loop process enables \model{} to progressively refine its focus, revisit uncertain moments, and allocate computation adaptively, leading to more efficient processing and reliable reasoning on long, complex videos.

We demonstrate the effectiveness and efficiency of \model{} by evaluating it on five long video understanding benchmarks, including MINERVA~\citep{nagrani2025minervaevaluatingcomplexvideo}, LVBench~\citep{xu2025lvbench}, Video-MME~\citep{fu2024videommefirstevercomprehensiveevaluation}, MLVU~\citep{zhou2024mlvu} and LongVideoBench~\citep{wu2024longvideobench}.
Compared to the existing agentic approaches, \model{} attains higher accuracy while using substantially less compute by formulating LVU as goal-conditioned observations.
Specifically, compared to the leading agentic method DeepVideoDiscovery (DVD)~\citep{zhang2025deepvideodiscoveryagentic}, \model{} achieves an average accuracy gain of \textbf{5.7\%}. 
What's more, on LVBench, \model{} achieves better performance while only consuming \textbf{18.4\%} inference time and \textbf{12.4\%} input tokens compared to DVD, validating the efficiency of \model{}.
We further conduct extensive ablation studies that highlight and justify the key design choices of \model{}.

%% file: sec/2_related_works.tex
\section{Related Work}
\label{sec:related_work}

\paragraph{Long Video Understanding}

The advancement in long video understanding (LVU) benchmarks~\citep{fu2024videommefirstevercomprehensiveevaluation, zhou2024mlvu, wu2024longvideobench, xu2025lvbench, chandrasegaran2024hourvideo} has extended video reasoning problem from short clips to realistic scenarios, involving multi-minute or hour-long videos.  
To address this, previous video-specific MLLMs \citep{lin2023video, zhang2024video, wang2023vamos,rana2024mvu, ryoo2025xgenmmvidblip3videoneed32, wang2024tarsierrecipestrainingevaluating} mainly focus on the challenge of excessive token inputs by extending the context length \citep{zhang2024long, chen2024longvilascalinglongcontextvisual}, reducing the video tokens \citep{wang2025adaretakeadaptiveredundancyreduction, shen2024longvu,li2024videochat, shu2024video} or  keyframe selection \citep{yu2023self, yao2025generative, tang2025adaptivekeyframesamplinglong, buch2022revisiting, wang2025videoitgmultimodalvideounderstanding,arnab2025temporalchainthoughtlongvideo,zou2025airenablingadaptiveiterative, tstar}.  
Notably, VAP~\citep{mavideo} also introduce the concept of ``action perception'' to LVU task, they treats key frame selection as
data acquisition in active perception and leverages a lightweight text-conditioned
video generation model to represent prior world knowledge. 
Instead, \model{} treats LVU as query-driven evidence seeking in video environments. As a result, \model{} tackles complex LVU task by focus perception in key regions, achieves significantly better efficiency. 

Recently, inspired by the great success of DeekSeek-R1~\citep{deepseekai2025deepseekr1incentivizingreasoningcapability}, several works ~\citep{feng2025video, wang2025videorftincentivizingvideoreasoning, wang-etal-2025-video, wang2025timezero} explore the Chain-of-thoughts video reasoning model. 
Later works~\citep{ge2025framemindframeinterleavedvideoreasoning, zhang2025thinkingvideosmultimodaltoolaugmented, tao2025mosschatvreinforcementlearningprocess, he2025framethinkerlearningthinklong, wang2025videothinkersparkingthinkingvideos, fu2025lover1advancinglongvideo, ouyang2025conanprogressivelearningreason,yuan2025videoexplorerthinkvideosagentic} explore the idea of ``Thinking with Video'', which incorporate visual CoT strategy to conduct coarse-to-fine video exploration. 
Compared to these methods, \model{} has two clear advantages: 
(1) query-adaptive, previous work mainly follows a coarse-to-fine schema with fixed FPS/resolution setup, instead, \model{} decides what/where/how to observe the video based on the query; 
(2) training-free, instead of generating large-scale training samples with reasoning trace, we directly employ an agentic approach and significantly reduce compute cost. 

\paragraph{Agentic Frameworks for Long Video Understanding}
LLM-based agents that combine reasoning, planning, and action have been widely studied~\cite{yao2023react,NEURIPS2023_1b44b878,erdogan2025planandact} in NLP domain, which also draws growing attention in LVU community.
To decouple the complex LVU task, early agentic frameworks \citep{kahatapitiya-etal-2025-language, park2025framesusefulefficientstrategies, wang2024videotree,zhang2023simple,zhang2025silvrsimplelanguagebasedvideo,ma2024drvideodocumentretrievalbased, wang2025videochata1thinkinglongvideos,fan2025agentickeyframesearchvideo,jeoung2024adaptivevideounderstandingagent} adopt a captioner–LLM design: video segments are converted into captions, which an LLM then uses the generated caption to answer the video query.
Meanwhile, several works~\citep{liu2025videomind, min2025morevqaexploringmodularreasoning, fan2024videoagentmemoryaugmentedmultimodalagent, menon2025caviarcriticaugmentedvideoagentic, kugo2025videomultiagentsmultiagentframeworkvideo,shi2025enhancingvideollmreasoningagentofthoughts,zhu2025activeo3empoweringmultimodallarge} utilize the idea of ``visual programming'', decompose the complex query into multiple steps to leverage expert modules. 
Reflection-based frameworks~\citep{chen2025lvagent, zhou2025reagentvrewarddrivenmultiagentframework} add a verification agent after the initial answering process to refine the reasoning. . 
Building on these works, recent studies \citep{shen2025vgentgraphbasedretrievalreasoningaugmentedgeneration, yan2025avaagenticvideoanalytics,zuo2025videolucydeepmemorybacktracking,zhang2025deepvideodiscoveryagentic, luo2024videoragvisuallyalignedretrievalaugmentedlong,chen2025lvagent, dong2025needqueryawarevisualintelligence, pang2025mrvideomapreduceprinciple} aim to improve evidence retrieval and reasoning efficiency in text space.
Notably, VGent~\cite{shen2025vgentgraphbasedretrievalreasoningaugmentedgeneration} constructs a caption-based graph to enable long-range retrieval and relational reasoning across segments.
VideoLucy~\cite{zuo2025videolucydeepmemorybacktracking}introduces a memory backtracking mechanism that allows the model to revisit earlier multi-scale text captions during multi-step reasoning.
Deep Video Discovery~\cite{zhang2025deepvideodiscoveryagentic} uses tool-based search to iteratively refine textual evidence over long videos.
Instead of relying on captioners, \model{} reasons directly over visual inputs through an iterative plan–observe–reflect process, selectively watching only what the query requires and maintaining a compact evidence record. 
This active, iterative video observation design preserves fine-grained grounding while avoiding the redundancy and overhead of caption-based LVU pipelines.

%% file: sec/3_method.tex
\section{Method}
\label{sec:method}

\begin{figure}[t]
  \centering
  \includegraphics[width=0.8\linewidth]{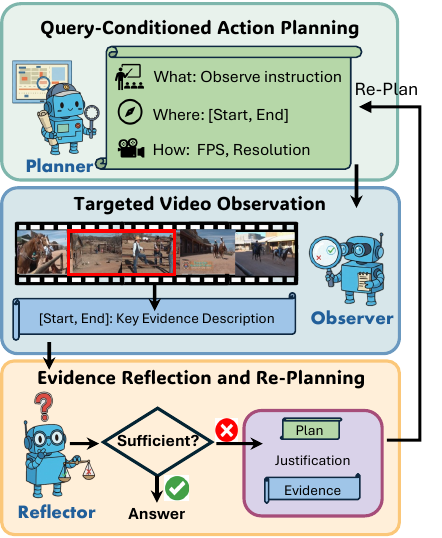}
  \caption{\textbf{Framework of Active Video Perception} (\model{}). \model{} operates by an iterative plan-observe-reflect process with MLLM agents. At each round, the planner decide what/where/how to interact with the video, the observer extract structured query-related evidence by executing the plan and the reflector evaluates the extracted evidence to decide whether an additional round is need.}
  \label{fig:method-overview}
  \vspace{-10pt}
\end{figure}

We present \textbf{Active Video Perception} (\model{}), an iterative evidence seeking framework for agentic LVU.  \model{} is inspired by the concept of \textit{active perception}~\citep{bajcsy2016revisitingactiveperception, bajcsy1988active, aloimonos2013active}, which argues ``a complete artificial agent necessarily must include the ability of knowing why it wishes to sense, and then choosing what to perceive, and determining how, when and where to achieve that perception''. 
Through the lens of active perception, we formulate LVU task as query-driven evidence seeking in video environments, where the LVU agent iteratively decides what, where and how to interact with the video to find the key evidence based on previous observation.

Concretely, as shown in \cref{fig:method-overview}, given a query $Q$ and a video $V$, \model{} runs an iterative plan-observe-reflect process with MLLM agents. 
In each round, a planner first proposes observation plan by choosing what to inspect, where to focus, and how to sample.
An observe agent executes that plan to extract compact, time-stamped evidence by observing the video purposefully.
A reflector verifies evidence against the query to estimate the confidence; if it exceeds the confidence threshold, \model{} outputs the answer and stops, otherwise it returns a justification to guide the next round of observation planning. 
We iterate this process until either a sufficiently confident answer is obtained or the round limit is reached. 
We introduce each component in detail as follow. 

\subsection{Query-Conditioned Action Planning}
\label{sec:3.1}

Inspired by active perception concept, instead of passively processing frames uniformly or converting into caption list, \model{} first plans deciding what, where, and how to observe the long video to obtain the query-related evidence. 
Specifically, \model{} leverage a planner (\textsc{Planner}) to decide what to look for, where to look, and how to observe in order to solve the give query.

\paragraph{Initial Plan.}
At round \(r{=}1\), given query \(Q\) and video \(V\), the \textsc{Planner} instantiates a concrete observation specification that states what to observe, the region to inspect, and how to sample. 
We initialize
\[
P^{(1)} \gets \textsc{Planner.Init}(Q),
\]
and represent it as
\[
P^{(1)}=\big(\texttt{what}^{(1)},\ \texttt{where}^{(1)},\ \texttt{how}^{(1)}\big).
\]
\begin{itemize}
  \item \textbf{\texttt{what}} is a brief, query-conditioned instruction naming the key evidence to seek (e.g., ``locate the moment the coach enters,'' ``determine who hands over the box,'' ``verify the scoreboard change''). 
  For complex query which requires multi-step reasoning, we prompt the \textsc{Planner} to first plan the initial observation and leave the following steps in the next rounds. 
  By decomposing the complex queries, \model{} achieves better handling in multi-hop reasoning and temporally dispersed evidence seeking. 
  \item \textbf{\texttt{where}} is a targeted temporal region \([t_s,t_e]\). 
It is seeded from: (i) explicit timestamps in \(Q\) (e.g., ``1{:}00–1{:}30''), (ii) soft textual cues (``opening scene,'' ``final minutes'').
When no prior is available, we first sweep the entire video at low cost (low \texttt{fps} and \texttt{spatial\_res}) to gather coarse evidence. 
Across rounds, this region can be tightened or shifted based on the Reflector’s feedback, enabling coarse-to-fine localization without dense scanning.
  \item \textbf{\texttt{how}} specifies sampling granularity for the long video observation, where \(\texttt{how}=(\texttt{fps},\ \texttt{spatial\_res})\). 
    The \textsc{Planner} determines the granularity of the targeted evidence and accordingly decides the sampling strategy. 
    By default, it adopts coarse settings (lower \texttt{fps} and \texttt{spatial\_res}) to perform low-cost exploration across the video and quickly identify potential evidence regions. 
    When finer details are required—such as subtle object interactions or small spatial cues, the \textsc{Planner} increases sampling density to ensure more precise perception.
    This adaptive design allows \model{} to allocate computation efficiently across granularities while maintaining high fidelity.
\end{itemize}

The resulting \(P\) serves as a compact, executable target observation that guides the Observer on what to looking for, which region of the video to inspect, and how to sample for efficient, query-focused observation.

\subsection{Targeted Video Observation}
Once the plan is generated, the observer (\textsc{Observer}, a MLLM) executes the plan to gather detailed, time-stamped evidence from the video. 
Specifically, in round $r$, given the plan \(P^{(r)}=(\texttt{what}^{(r)},\,\texttt{where}^{(r)},\,\texttt{how}^{(r)})\), the \textsc{Observer} inputs the the query \(Q\) and instruction in $\texttt{what}^{(r)}$, the video segment defined by the temporal \texttt{where} and uses the sampling strategy in \texttt{how} (fps and spatial resolution). 
Instead of generating free-form text responses, the \textsc{Observer} is prompted to produce structured, timestamp-aware evidence text in the form of
\[
E \in \{([\texttt{start}_i,\ \texttt{end}_i],\, d_i)\}_{i=1}^{N},
\]
where each \(d_i\) is a concise, query-conditioned description of the visual event within the time interval \([\texttt{start}_i,\ \texttt{end}_i]\). 
Specifically, we maintain an evidence list \(\mathcal{E}\) that accumulates evidence across rounds.
At each round \(r\), the \textsc{Observer} generates new evidence $E^{r}$ and append it to the cumulative evidence list:
\[
E^{r} = \textsc{Observer}(V,\, Q,\, P^{(r)}),\quad \mathcal{E}  \leftarrow \mathcal{E}  \cup E^{r}.
\]
This cumulative evidence list \(\mathcal{E}\) serves as the working memory of \model{}, allowing the reflector to assess sufficiency based on all past evidence and guiding the \textsc{Planner}’s subsequent updates.
Compared with free-form captioning, this design yields more stable, query-relevant evidence and leads to better grounded reasoning over long videos.
This targeted video observation design allows \model{} to perceive only the most query-relevant portions of the video, keeping it efficient and avoiding redundant or irrelevant information.

\begin{algorithm*}[t]
\caption{\textbf{Active Video Perception}(\model{})}
\label{alg:avp_min_history}
\small
\DontPrintSemicolon
\RestyleAlgo{plain}
\SetKwInOut{Input}{Inputs}
\SetKwInOut{Output}{Output}

\Input{Video $V$,\quad Query $Q$,\quad Max Rounds $R_{\max}$,\quad Confidence Threshold $\tau_{\mathrm{conf}}$}
\Output{Answer $A$,\quad Justification $J$,\quad Evidence List $\mathcal{E}$,\quad History $H$}

$P^{(1)} \gets \textsc{Planner.Init}(Q)$;\quad $ \mathcal{E} \gets [\,]$; \quad $H \gets [\,]$ \tcp*{Init plan, evidence list and history}
\For{$r \gets 1$ \KwTo $R_{\max}$}{
  $E^{(r)} \gets \textsc{Observer}(V, Q, P^{(r)})$\;
  $\mathcal{E} \gets \mathcal{E} \cup E^{(r)}$ \tcp*{accumulate this round’s evidence}

  $(C^{(r)}, J^{(r)}) \gets \textsc{Reflector}(Q, \mathcal{E})$ \tcp*{evidence reflection: confidence \& justification}

  \If{$C^{(r)} \ge \tau_{\mathrm{conf}}$}{
     $A \gets \textsc{Reflector.ExtractAnswer}(J^{(r)})$ \tcp*{answer is entailed by justification}
     \Return $A,\, J^{(r)},\, \mathcal{E},\, H$\;
  }

  \If{$r = R_{\max}$}{
     $A \gets \textsc{Reflector.ForceAnswer}(Q, \mathcal{E})$ \tcp*{force to give answer on final round}
     \Return $A,\, J^{(r)},\, \mathcal{E},\, H$\;
  }

  $H \gets H \cup \{(P^{(r)},\, E^{(r)},\, J^{(r)})\}$ \tcp*{append plan  \& evidence \& justification to history}
  $P^{(r+1)} \gets \textsc{Planner.REPLAN}(Q, H, J^{(r)})$ \tcp*{re-plan for additional observation}
}
\end{algorithm*}

\subsection{Evidence Reflection and Re-Planning}
\label{subsec:method}
After each observation round, \model{} employs a reflector (\textsc{Reflector}) to evaluate the sufficiency of the accumulated evidence and decide whether additional observation is required. 
The \textsc{Reflector} verifies how well the collected evidence supports an answer, and when confidence is insufficient, it provides feedback for the next round of planning.

\paragraph{Evidence Reflection.}
At round $r$, given the query \(Q\) and the current cumulative evidence list \(\mathcal{E}\), the \textsc{Reflector} jointly produces a query confidence score \(C^{(r)}\)and a justification \(J^{(r)}\):
\[
(C^{(r)},\, J^{(r)}) = \textsc{Reflector}(Q,\, \mathcal{E}),
\]
where \(C^{(r)} \in [0,1]\) measures the confidence in evidence sufficiency to answer the given query, and \(J^{(r)}\) specifies which answer the current evidence supports or what information is still missing.
If the confidence is higher than the confidence threshold \(\tau_{\text{conf}}\), the \textsc{Reflector} directly extracts the final answer from \(J^{(r)}\); otherwise, the justification highlights missing or uncertain cues to guide the next round of planning step.

\paragraph{History Update and Re-Planning.}
When confidence remains below the threshold, the Reflector appends the current observation and justification to the running history $H$. 
The history provides the \textsc{Planner} with a concise summary of what has been inspected, verified, or left unresolved. 
The \textsc{Planner} then refines its next plan using this feedback:
\[
P^{(r+1)} = \textsc{Planner.REPLAN}(Q,\, H,\, J^{(r)}),
\]
shifting attention toward the regions, entities, or temporal spans identified as uncertain by the Reflector.

By iteratively running the plan-observe-reflect process, \model{} forms a closed-loop perception–reasoning cycle that continuously refines its focus until the gathered evidence becomes sufficient. 
This iterative design allows the system to adaptively reason over long videos, reducing computation, and maintaining grounded, query-aligned understanding.
We present full algorithm in \cref{alg:avp_min_history}. 

%% file: sec/4_exp_setup.tex
\section{Experimental Setup}
\label{sec:expsetup}

\subsection{Datasets}
We evaluate \model{} on five diverse long video understanding benchmarks: 

\noindent(1) \textbf{MINERVA}~\citep{nagrani2025minervaevaluatingcomplexvideo} is a recent challenging video reasoning benchmark consisting of $1515$ hand-crafted questions. The average video duration is $12$ minutes. 

\noindent(2) \textbf{LVBench}~\citep{xu2025lvbench} is a benchmark specifically designed for long video understanding which includes $1549$ multiple-choice questions across 103 hour-long videos.

\noindent(3) \textbf{MLVU}~\citep{zhou2024mlvu} is a multi-task Long Video Understanding Benchmark for the comprehensive and in-depth evaluation of LVU.
We use the multiple-choice QA samples from the MLVU test split, containing $2175$ video QA samples with more than 15 minutes average video duration.

\noindent(4) \textbf{Video-MME}~\citep{fu2024videommefirstevercomprehensiveevaluation} is a comprehensive evaluation benchmark for video analysis from short to long videos (average min for long split video).
We use the standard split of Video-MME, which contains $2700$ samples designed for both perception and reasoning tasks ($900$ samples with $41$min average duration for the long split).

\noindent(5) \textbf{LongVideoBench (LVB)}~\citep{wu2024longvideobench} is a video QA benchmark that highlights referred reasoning questions, which are dependent on long frame inputs. 
We test on the public validation split, which contains $1337$ video reasoning questions ($533$ samples with 15-60 min video for long split). 

\subsection{Evaluation Metrics}
We evaluate \model{} under the multiple-choice QA setting. 
We use standard accuracy metrics for all experiments. 
We do not include auxiliary subtitle for all benchmarks.  

\subsection{Implementation Details}
We adopt Gemini-2.5-Pro\footnote{2025-06-17 version} \citep{comanici2025gemini25pushingfrontier} as our default MLLM agent for all components. 
We also provide the results with lightweight Gemini-2.5-Flash model in \cref{tab:main_table} and \cref{tab:module_ablation}. 
We provided more ablation with different backbone models (including open-source models) in appendix. 
For fair comparison, we fix the max input token as $128K$. 
If the input video (region) exceeds this budget, we uniformly sample the max frames that within the token limit. 
For spatial token setup (\texttt{spatial\_res}), we follow Gemini's MediaResolution setup to have 2 scales (low, medium), where low and medium are $66$ and $258$ tokens per frame, respectively. 
We set the max rounds $R_{\max}$ as $3$ and confidence threshold $\tau_{\mathrm{conf}}$ as $0.7$. We provide additional analysis for the design choices in \cref{subsec:abla}. 
We provided more implementation details (including detailed prompts) and analysis in appendix. 

%% file: sec/5_results.tex
\section{Results}

\begin{table*}[t]
  \centering
  \setlength{\tabcolsep}{10pt}
\resizebox{2\columnwidth}{!}{
\begin{tabular}{l c c c cc cc}
\toprule
\multirow{2}{*}{Methods}  &
MINERVA &
LVBench &
MLVU &
\multicolumn{2}{c}{Video-MME} &
\multicolumn{2}{c}{LongVideoBench} \\
\cmidrule(lr){2-2}\cmidrule(lr){3-3}\cmidrule(lr){4-4}\cmidrule(lr){5-6} \cmidrule(lr){7-8}
 & Overall & Overall & Test & Overall & Long & Val & Long \\
\midrule
\rowcolor{gray!15}\multicolumn{8}{l}{\textit{General-Purpose MLLMs}} \\
Seed-1.5-VL~\citep{guo2025seed15vltechnicalreport}        & -    & 64.6 & \underline{82.1} & 77.9 & - & 74.4 & - \\
Qwen-3-VL~\citep{qwen3vl2025}  & -    & 67.7 & \textbf{84.3} & 79.2 & - & - & - \\
GPT-4o~\citep{openai2024gpt4ocard}             & 45.5 & 48.9 & 54.9 & 71.9 & 65.3 &66.7 & 60.9\\
GPT-4.1~\citep{openai_gpt4.1_2025}            & 54.0 & 63.4 & -    & 72.0 & - & - & - \\
Gemini-2.5-Flash~\citep{comanici2025gemini25pushingfrontier}   & 54.6 & 56.7 & 72.4 & 74.2 & 69.1 & 66.2 & 61.8\\
Gemini-2.5-Pro~\citep{comanici2025gemini25pushingfrontier}     & \underline{61.8} & 67.4 & 79.6 & \underline{82.4} & 77.6 & 69.8 & 66.6\\
\midrule
\rowcolor{gray!15}\multicolumn{8}{l}{\textit{Video-Specific MLLMs}} \\
LongVU~\citep{shen2024longvu}             & -    & -    & 65.4 & 60.6 & 59.5 & - & - \\
AdaReTaKe~\citep{wang2025adaretakeadaptiveredundancyreduction}          & -    & 53.3 & 78.1 & 73.5 & 65.0  & 67.0 & -\\
Video-RTS~\citep{wang-etal-2025-video}          & 37.8 & 43.2 & -    & 63.0 & 54.1 &  56.6 & 52.2\\ 
FrameMind~\citep{ge2025framemindframeinterleavedvideoreasoning}          & -    & -    & 48.6 & 60.9 & 57.5 & - & - \\ 
\midrule
\rowcolor{gray!15}\multicolumn{8}{l}{\textit{Agentic Video Frameworks}} \\
VideoAgent~\citep{wang2024videoagentlongformvideounderstanding}         & -    & 29.3 & 64.4 & -    & 46.4 & - & -\\
VideoTree~\citep{wang2024videotree}          & 40.2 & 28.8 & 60.4 & 60.6 & 54.2 & - & -\\
SiLVR~\citep{zhang2025silvrsimplelanguagebasedvideo}              & 44.4 & -    & 45.2 & 74.1 & \underline{77.7} & - &  -\\
VideoLucy~\citep{zuo2025videolucydeepmemorybacktracking}          & -    & 58.8 & 76.1 & 72.5 & 66.8 & - & - \\
Vgent~\citep{shen2025vgentgraphbasedretrievalreasoningaugmentedgeneration}           & -    & -    & 72.1 & 68.9 & - & 59.7 & - \\
LVAgent~\citep{chen2025lvagent}            & -    & -    & {\textcolor{lightgray}{83.9}} & \underline{81.7} & 74.3 & {\textcolor{lightgray}{80.0}} & - \\
DeepVideoDiscovery (DVD)~\citep{zhang2025deepvideodiscoveryagentic} & - & \underline{74.2} & -    & -    & 67.3 & \underline{71.6} & \underline{68.6}  \\
\midrule
\rowcolor{gray!15}\multicolumn{8}{l}{\textit{Active Video Perception (Ours)}} \\

\model{} w Gemini-2.5-Flash
  & 56.9 {\scriptsize\textbf{\textcolor{blue}{(+2.3)}}}
  & 63.8 {\scriptsize\textbf{\textcolor{blue}{(+7.1)}}}
  & 74.1 {\scriptsize\textbf{\textcolor{blue}{(+1.7)}}}
  & 81.2 {\scriptsize\textbf{\textcolor{blue}{(+7.0)}}}
  & 76.7 {\scriptsize\textbf{\textcolor{blue}{(+7.6)}}}
  & 70.2 {\scriptsize\textbf{\textcolor{blue}{(+4.0)}}}
  & 65.5 {\scriptsize\textbf{\textcolor{blue}{(+3.7)}}} \\

\model{} w Gemini-2.5-Pro
  & \textbf{65.6 {\scriptsize\textbf{\textcolor{blue}{(+3.8)}}}}
  & \textbf{74.8 {\scriptsize\textbf{\textcolor{blue}{(+7.4)}}}}
  & \textbf{84.3 {\scriptsize\textcolor{blue}{(+4.7)}}}
  & \textbf{85.3 {\scriptsize\textcolor{blue}{(+2.9)}}}
  & \textbf{81.9 {\scriptsize\textcolor{blue}{(+4.3)}}}
  & \textbf{73.4 {\scriptsize\textcolor{blue}{(+3.6)}}}
  & \textbf{70.0 {\scriptsize\textcolor{blue}{(+3.4)}}} \\

\bottomrule
\end{tabular}}
\vspace{-0.3em}
\caption{Comparison with general-purpose MLLMs, Video-specific MLLMs, and agentic video frameworks on five long video understanding benchmarks (MINERVA, LVBench, MLVU, Video-MME, LongVideoBench). 
We \textbf{bold} the best and \underline{underline} the second-best result in each column. 
Results shows that \model{} achieves best overall accuracy on all datasets across different baselines, achieving significant improvements on its backbone model (in \textcolor{blue}{blue}) across all benchmark. 
We \textcolor{lightgray}{gray} out the results that use auxiliary subtitle information.}
\label{tab:main_table}
\end{table*}

\subsection{Main Results on Long Video Benchmarks}

\cref{tab:main_table} presents a comprehensive comparison of \model{} against existing general-purpose MLLMs ~\citep{guo2025seed15vltechnicalreport, openai2024gpt4ocard, openai_gpt4.1_2025, comanici2025gemini25pushingfrontier}, video-specific MLLMs~\citep{shen2024longvu, wang2025adaretakeadaptiveredundancyreduction, wang-etal-2025-video, ge2025framemindframeinterleavedvideoreasoning}, and agentic video frameworks \citep{wang2024videoagentlongformvideounderstanding, wang2024videotree, zhang2025silvrsimplelanguagebasedvideo, zuo2025videolucydeepmemorybacktracking, shen2025vgentgraphbasedretrievalreasoningaugmentedgeneration, chen2025lvagent, zhang2025deepvideodiscoveryagentic} across five video understanding benchmarks: MINERVA~\citep{nagrani2025minervaevaluatingcomplexvideo}, LVBench~\citep{xu2025lvbench}, MLVU~\citep{zhou2024mlvu}, Video-MME~\citep{fu2024videommefirstevercomprehensiveevaluation} and LongVideoBench~\citep{wu2024longvideobench}. 

\paragraph{Comparison with MLLMs.}
Among general-purpose multimodal LLMs, proprietary systems such as Gemini-2.5-Pro \citep{comanici2025gemini25pushingfrontier} and Seed-1.5-VL \citep{guo2025seed15vltechnicalreport} achieve strong overall results but still fall short of our proposed \model{}.
In particular, \model{} (w/ Gemini-2.5-Pro) surpasses the state-of-the-art Gemini-2.5-Pro model~\citep{comanici2025gemini25pushingfrontier} by $\textbf{4.5\%}$ average accuracy over all benchmarks, demonstrating that direct inference over full length remains insufficient for complex, long-horizon queries that require targeted evidence seeking.
\model{} (w/ Gemini-2.5-Flash) also outperforms its backbone by $\textbf{4.4\%}$, showing generalization ability of the proposed framework in weaker backbone MLLMs. 
Meanwhile, \model{} significantly outperforms the video-specific MLLMs, including compression-based methods~\citep{shen2024longvu, wang2025adaretakeadaptiveredundancyreduction} and (visual) Chain-of-Thoughts methods ~\citep{wang-etal-2025-video, ge2025framemindframeinterleavedvideoreasoning}. 
This result highlights the active perception concept for long video understanding and encourages future research.

\paragraph{Comparison with Agentic Frameworks.}
Within the class of agentic video reasoning systems, \model{} consistently achieves the best (or second-best) results across all benchmarks. 
We compare \model{} with six recent agentic video frameworks, including VideoAgent~\citep{wang2024videoagentlongformvideounderstanding}, VideoTree~\citep{wang2024videotree}, SiLVR~\citep{zhang2025silvrsimplelanguagebasedvideo}, VideoLucy~\citep{zuo2025videolucydeepmemorybacktracking}, LVAgent~\citep{chen2025lvagent} and DeepVideoDiscovery (DVD)~\citep{zhang2025deepvideodiscoveryagentic}. 
We find that \model{} achieves best performance against all baseline methods and significant improvement compared to the backbone model in all benchmarks.  
Comparing to the recent VideoLucy and DVD methods, \model{} achieves $\textbf{10.5\%}$ and $\textbf{5.7\%}$ average improvements while both using strong LLM backbones (DeepSeek-R1~\citep{deepseekai2025deepseekr1incentivizingreasoningcapability} for VideoLucy, and OpenAI-o3~\citep{openai_o3_2025_systemcard} for DVD). 
We also compared the efficiency in term of inference time with DVD in \cref{tab:efficiency_breakdown}, showing \model{} is not only more performant, but also significantly efficient. 
These results validate the effectiveness of active perception for long video understanding
: rather than passively encoding frames, \model{} plans what to observe, observes purposefully, and reflects adaptively, leading to higher accuracy and greater efficiency than both MLLMs and recent agentic frameworks.

\subsection{Quantitative Analysis}
In this section, we analyze different aspect of \model{}, including efficiency analysis, ablation study on \model{}'s design choices. We provided more quantitative analysis in the appendix.

\subsubsection{Efficiency Analysis}
As shown in \cref{tab:efficiency_breakdown}, we evaluate inference efficiency on LVBench in terms of average runtime, average input token count, and accuracy.
DVD~\citep{zhang2025deepvideodiscoveryagentic} requires $790.5$s per video and processes on average $1.07$M tokens. 
Notably, a finer breakdown shows that its captioning stage alone takes $637.2$s and consumes roughly $0.9$M tokens.
In contrast, \model{} eliminates this query-agnostic captioning stage and performs only targeted query reasoning, reducing inference time to $145.3$s, achieving 5.44$\times$ faster (\textbf{81.6\%} reduction). 
Meanwhile, \model{} only consumes $\textbf{12.4\%}$ of the input tokens compared to DVD while improving the LVBench accuracy.
These results indicate that \textbf{actively} deciding what, where, and how to observe not only removes redundant caption processing but also strengthens reasoning by concentrating computation on query-relevant content.

\begin{table}[t]
  \centering
  \resizebox{1\columnwidth}{!}{
  \begin{tabular}{lccc}
    \toprule
    \textbf{Method} & \textbf{Avg. Inference Time (s)} & \textbf{Avg. Input Tokens (K)} & \textbf{Acc} \\
    \midrule
    DVD              & 790.5 &            1071.6           & 74.2 \\
    \model{} (Ours)  & \textbf{145.3} &      \textbf{132.5}      & \textbf{74.8} \\
    \bottomrule
  \end{tabular}
  }
  \vspace{-0.35em}
  \caption{Efficiency comparison on LVBench. 
  We report average inference time in seconds, average input token count, and accuracy. 
  By actively querying the video rather than passively captioning all clips, \model{} achieves better overall efficiency and accuracy.}
  \label{tab:efficiency_breakdown}
\end{table}

\begin{figure*}[t]
  \centering
  \includegraphics[width=0.97\linewidth]{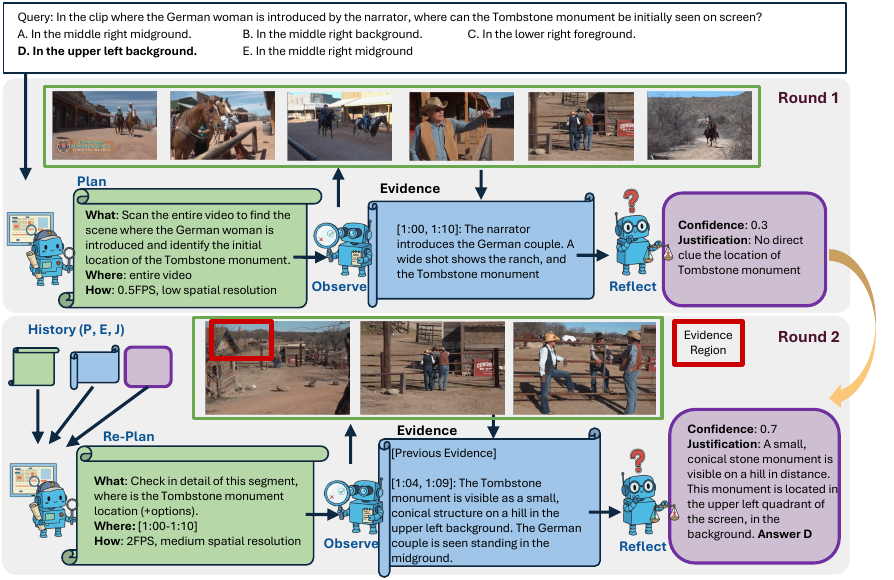}
\caption{\textbf{Qualitative example of \model{}}. 
Given a multiple-choice query about the Tombstone monument's first on-screen
appearance, Round~1 performs a coarse scan of the entire video
(0.5~FPS, low resolution) and localizes a candidate interval
[1{:}00, 1{:}10], but the \textsc{Reflector} judges the evidence insufficient.
Round~2 re-plans a targeted pass over this window (2~FPS, medium resolution),
enabling the \textsc{Observer} to localize the monument in the upper-left background
and the \textsc{Reflector} to confidently select the correct answer (option~D) and halt.}

\label{fig:visualization}
\end{figure*}

\subsubsection{Ablation Study}
\label{subsec:abla}
\paragraph{\model{} Components.}
We conduct a step-wise ablation to assess the contribution of each component in \model{}. As shown in \cref{tab:component_ablation}, introducing the \textsc{Planner} notably improves both MINERVA and LVBench accuracy, demonstrating the benefit of query-conditioned multi-step exploration over static observation. 
The \textsc{Planner} guides the agent to allocate computation toward potentially informative regions rather than processing frames uniformly.
Adding the \textsc{Reflector} yields a further performance gain, confirming that iterative process enhances reasoning trustworthy.  
Together, these results highlight that active perception, planning what to observe and reflecting on what has been seen substantially strengthens long video understanding.

\begin{table}[t]
  \centering
  \resizebox{1\columnwidth}{!}{
  \begin{tabular}{lcc}
    \toprule
    \textbf{Method} & \textbf{MINERVA} & \textbf{LVBench} \\
    \midrule
    Observer (Baseline) & 60.8  & 67.4 \\
    Planner + Observer & 63.9 & 72.6 \\
    Planner + Observer + Reflector (\model{}) & \textbf{65.6} & \textbf{74.8} \\
    \bottomrule
  \end{tabular}
  }
  \vspace{-0.35em}
\caption{\textbf{Component ablation of \model{}.} Adding the Planner and then the Reflector on top of the Observer baseline consistently improves MINERVA and LVBench accuracy, showing that query-conditioned planning and reflection are key to \model{}’s performance.}
  \label{tab:component_ablation}
\end{table}

\paragraph{Model Selection.}

\Cref{tab:module_ablation} examines the impact of varying the model selection across \textsc{}{Planner}, \textsc{Observer}, and \textsc{Reflector} within \model{} under Gemini-2.5~\citep{comanici2025gemini25pushingfrontier} family (we add additional model ablation in supp.). 
We observe that both benchmarks benefit from stronger components, but their sensitivities differ.
Across both benchmarks, the largest gains consistently come from a stronger Observer.
This suggests that reliable evidence acquisition is the primary bottleneck, even for complex multi-hop reasoning queries, as effective planning and reflection depend on first
locating and gathering the relevant visual evidence. The best configuration uses stronger models across all three modules, confirming that \model{}'s active perception design benefits from the synergy between observation, planning, and reflection.

\begin{table}[t]
  \centering
  \resizebox{0.95\linewidth}{!}{
  \setlength{\tabcolsep}{5pt}
  \begin{tabular}{ccccc}
    \toprule
    \textbf{\textsc{Planner}} & \textbf{\textsc{Observer}} & \textbf{\textsc{Reflector}} & \textbf{MINERVA} & \textbf{LVBench} \\
    \midrule
    2.5-Flash & 2.5-Flash & 2.5-Flash & 56.9 & 63.8 \\
    2.5-Pro   & 2.5-Flash & 2.5-Pro   & 60.2 & 67.6 \\
    2.5-Flash & 2.5-Pro   & 2.5-Flash & 63.6 & 71.8 \\
    2.5-Pro   & 2.5-Pro   & 2.5-Pro   & \textbf{65.6} & \textbf{74.8} \\

    \bottomrule
  \end{tabular}
  }
  \vspace{-0.35em}
\caption{\textbf{Agent MLLM selection within \model{}.} We vary Gemini-2.5 Flash/Pro backbones for the \textsc{Planner}, \textsc{Observer}, and \textsc{Reflector}, stronger components consistently improve performance on both benchmarks.}
  \label{tab:module_ablation}
\end{table}

\vspace{-10pt}
\paragraph{Max Round Limit.}
\Cref{tab:iteration_ablation} studies how the number of Plan–Observe–Reflect rounds affects performance.
Both MINERVA and LVBench show steady gains from one to three rounds, confirming that iterative reasoning enables \model{} to progressively refine its evidence set and improve decision confidence.
The improvement is more pronounced on MINERVA, where multi-hop reasoning benefits from repeated reflection and targeted re-observation.
Beyond three rounds, performance saturates, suggesting that \model{} has already acquired sufficient evidence and additional cycles bring limited benefit.
This result validates the efficiency of our design, \model{} achieves strong reasoning capability with only a few lightweight interaction rounds.

\begin{table}[t]
  \centering
  \resizebox{0.7\linewidth}{!}{
  \setlength{\tabcolsep}{10pt}
  \begin{tabular}{ccc}
    \toprule
    \textbf{Max Rounds} & \textbf{MINERVA} & \textbf{LVBench} \\
    \midrule
    1  & 63.9 & 72.6 \\
    2  & 65.0 & 74.6 \\
    3  & \textbf{65.6} & \textbf{74.8} \\
    5  & 65.5 & 74.6 \\
    \bottomrule
  \end{tabular}
  }
  \vspace{-0.4em}
  \caption{\textbf{Ablation on max round limit.} Increasing the number of max round limit improves performance on both benchmarks and gets best results by three rounds, indicating that only a few interaction steps are sufficient. }
  \label{tab:iteration_ablation}
\end{table}

\subsection{Visualization}
In \cref{fig:visualization}, we illustrate how \model{} acquires and verifies evidence through a multi-round Plan–Observe–Reflect loop on a long video.
Given the query, “In the clip where the German woman is introduced by the narrator, where can the Tombstone monument be initially seen on screen?”, Round 1 uses a coarse, uniform sweep to localize candidate moments (0.5 FPS, low resolution). 
This pass narrows the search to the [1:00, 1:10] interval but the reflector flags the observations as insufficient due to lack of detail, prompting a refined follow-up (Round 2).
In Round 2, the planner schedules a targeted revisit over [1:00, 1:10] at 2 FPS with medium resolution, and the observer extracts query-relevant cues: the Tombstone monument appears as a small, conical structure on a hill in the upper-left background while the German couple stands in the mid-ground. The evidence list is now sufficient for the reflector to stop and produce the final answer, demonstrating \model{}’s coarse-to-fine scheduling, evidence-grounded verification.
We provided additional visualization samples with different scenario (start with a grounded video region from query prior and refine the region based on the observation in the next round) and failure case in appendix.

%% file: sec/6_conclusion.tex
\section{Conclusion}
\label{sec:conclusion}
Inspired by active perception theory, we present \textbf{Active Video Perception} (\model{}), which handles long video understanding as an iterative, query-driven evidence seeking process.
Rather than passively caption the video frames, \model{} treats the video as an interactive environment and actively decides what to inspect, where to focus, and at what granularity in order to acquire compact, time-stamped evidence directly from pixels.
Concretely, \model{} runs an iterative plan–observe–reflect process using MLLM agents.
Empirically, \model{} achieves best overall accuracy among agentic frameworks across five long video benchmarks, and surpasses the leading agentic method (DVD) by \textbf{5.7\%} in average accuracy while only requiring \textbf{18.4\%} inference time and \textbf{12.4\%} input tokens.
Our ablation study shows that \model{} achieves significant improvement under different MLLM backbones, validating the robustness. 
Looking ahead, an exciting direction is extending active video perception to embodied agents that must decide what and when to observe while acting under real-world physical constraints.

%% file: sec/X_suppl.tex
\clearpage
\appendix
\section*{Appendix}

In appendix, we present the following: limitations (\cref{sec:appendix_limitation}), additional quantitative results and analysis (\cref{sec:appendix_quantitative}), additional qualitative analysis (\cref{sec:appenddix_qualitative}),
additional implementation details (\cref{sec:appendix_implement}).

\section{Limitations}
\label{sec:appendix_limitation}
While \model{} achieves strong performance and efficiency gains across multiple LVU benchmarks, it also has several practical limitations that point to promising future work rather than fundamental constraints.

First, we primarily evaluate \model{} in the standard offline video QA setting, where the full video is available. 
An exciting direction for future work is to explore how the same active evidence-seeking framework operates in broader scenarios, such as embodied or online streaming environments where an agent must perceive and act in real time.
Second, \model{} currently uses prompting to drive planning and observation, learning policies that optimize long horizon sensing efficiency under resource and latency constraints (e.g., via reinforcement learning or differentiable planners) would be a complementary direction that builds on the same architecture.

\section{Additional Quantitative Results and Analysis}
\label{sec:appendix_quantitative}
\subsection{Reasoning Trace Analysis}
Proposed by MINERVA \citep{nagrani2025minervaevaluatingcomplexvideo}, the MiRA (MINERVA Reasoning Assessment) score is a reference-based, LLM-as-a-judge metric for evaluating the quality of multimodal models' step-by-step reasoning traces for video question answering. It assesses a model's generated reasoning against a ground-truth trace using the four axes of the MINERVA rubric: Perceptual Correctness, Temporal Localization, Logical Reasoning, and Completeness. This normalized score helps analyze why models succeed or fail beyond just the final answer's accuracy, specifically highlighting weaknesses in video-centric aspects like temporal grounding and perception.

As shown in \cref{tab:mira_single_col}, \model{} achieves the highest overall MiRA score, outperforming all baselines across key reasoning dimensions. 
Compared to single-pass MLLMs, \model{} delivers substantially stronger temporal localization, logical reasoning, and correctness.
These improvements indicate that actively collecting structured, query-conditioned evidence leads to higher-quality reasoning traces besides higher final accuracy. 
In particular, \model{}’s gains in temporal and completeness highlight the benefit of iterative planning and reflection for complex multi-hop queries.

\begin{table}[t]
  \centering
  \setlength{\tabcolsep}{6pt}
  \resizebox{\columnwidth}{!}{
  \begin{tabular}{l c c c c c c}
    \toprule
    \multirow{2}{*}{\textbf{Method}} & \multirow{2}{*}{\textbf{Acc.\ \%}} & \multicolumn{5}{c}{\textbf{MiRA Score} $\uparrow$} \\
    \cmidrule(lr){3-7}
     &  & \textbf{P} & \textbf{T} & \textbf{L} & \textbf{C} & \textbf{Total} \\
    \midrule
    OpenAI o1           & 43.5 & 0.52 & 0.52 & \textit{0.86} & \textit{0.88} & 0.69 \\
    GPT-4o                & 45.5 & 0.57 & 0.67 & 0.77 & 0.79 & 0.70 \\
    Gemini 2.0 Flash    & 53.5 & \textbf{0.62} & \textit{0.75} & 0.83 & 0.82 & \textit{0.75} \\
    Gemini 2.5 Pro & \textit{61.8} & \textit{0.60} & 0.62 & \textbf{0.97} & 0.78 & 0.74 \\
    \midrule
    \model{} (Ours) & \textbf{65.6} & \textbf{0.62} & \textbf{0.82} & \textbf{0.97} & \textbf{0.93} & \textbf{0.84} \\
    \bottomrule
  \end{tabular}}
  \vspace{-0.4em}
  \caption{Reasoning trace quality check on MINERVA. We report multiple-choice accuracy and MiRA scores normalized to be between 0 and 1. P: Perceptual Correctness, T: Temporal Localization: L: Logical Reasoning: C: Completeness. The best result is in \textbf{bold}, and the second best is in \textit{italic}.}
  \label{tab:mira_single_col}
\end{table}

\subsection{Full Results for LVBench}

As shown in \cref{tab:lvbench_categories}, \model{} achieves the best overall accuracy on LVBench, outperforming all prior systems including the strongest agentic baseline DVD~\citep{zhang2025deepvideodiscoveryagentic}. The gains are most pronounced on splits that require integrating information over long temporal ranges: \model{} delivers the highest scores on Event Understanding, Temporal Grounding, and Summarization, indicating that its plan–observe–reflect loop is effective at steering perception toward query-relevant moments and aggregating evidence across distant segments. On Key Information Retrieval, Entity Recognition, and Reasoning, \model{} remains competitive with DVD, while still substantially outperforming powerful generic MLLMs across all question types. 
These results suggest that explicit active video perception is crucial for long video understanding.

\begin{table}[t]
  \centering
  \resizebox{\linewidth}{!}{
  \begin{tabular}{lccccccc}
    \toprule
    \textbf{Methods} & \textbf{ER} & \textbf{EU} & \textbf{KIR} & \textbf{TG} & \textbf{Rea} & \textbf{Sum} & \textbf{Overall} \\
    \midrule
    GPT-4o             & 48.9 & 49.5 & 48.1 & 40.9 & 50.3 & 50.0 & 48.9 \\
    OpenAI o3          & 57.6 & 56.4 & 62.9 & 46.8 & 50.8 & 67.2 & 57.1 \\
    AdaReTAKe         & 53.0 & 50.7 & 62.2 & 45.5 & 54.7 & 37.9 & 53.3 \\
    VideoTree          & 30.3 & 25.1 & 26.5 & 27.7 & 31.9 & 25.5 & 28.8 \\
    VideoAgent         & 28.0 & 30.3 & 28.0 & 29.3 & 28.0 & 36.4 & 29.3 \\
    VCA                & 43.7 & 40.7 & 37.8 & 38.0 & 46.2 & 27.3 & 41.3 \\
    MR. Video          & 59.8 & 57.4 & 71.4 & 58.8 & 57.7 & 50.0 & 60.8 \\
    DVD          & \textbf{73.4} & \textit{73.3} & \textbf{80.4} & \textit{72.3} & \textbf{70.7} & \textit{74.1} & \textit{74.2}\\
    \midrule
    \textbf{\model{} (Ours)} & \textit{71.9} & \textbf{76.7} & \textit{80.1} & \textbf{73.6} & \textit{67.7} & \textbf{75.9} & \textbf{74.8} \\
    \bottomrule
  \end{tabular}
  }
  \vspace{-0.4em}
\caption{\textbf{Results by question type on LVBench.}
We report performance across six official LVBench splits:
\textit{Entity Recognition (ER)},
\textit{Event Understanding (EU)}, 
\textit{Key Information Retrieval (KIR)}, 
\textit{Temporal Grounding (TG)},  
\textit{Reasoning (Rea)},
and \textit{Summarization (Sum)}. 
Accuracy (\%) is computed as Correct / Total for each split. }

\label{tab:lvbench_categories}
\end{table}

\subsection{Additional Ablation Study}

\noindent\textbf{Controlled Backbone Comparison.}
In ~\cref{tab:main_table}, we follow prior works (VideoTree, DVD) and report the best performance of each method.
To further ensure a fair comparison, we extend same-backbone evaluations (w/ OpenAI-o3) to additional agentic methods on the Video-MME long split.
As shown in ~\cref{tab:fair_comparison}, \model{} achieves the highest overall accuracy and the lowest inference time among all compared methods, demonstrating improvements in both effectiveness and efficiency under a controlled backbone setting.

\begin{table}[h]
\centering
\resizebox{1\columnwidth}{!}{
\begin{tabular}{lcccc}
\toprule
 & VideoTree & SiLVR & DVD & \textbf{AVP} \\
\midrule
Long Split Acc.\ (\%) $\uparrow$ & 61.2 & 66.8 & 67.3 & \textbf{76.8} \\
Inference Time (s) $\downarrow$ & 145.4 & 442.2 & 612.8 & \textbf{102.3} \\
\bottomrule
\end{tabular}
}
\caption{Controlled backbone comparison (OpenAI-o3) with agentic methods on Video-MME long split. We report long split accuracy (\%) and average inference time (s).}
\label{tab:fair_comparison}
\end{table}



\paragraph{Different Backbone MLLM Selection within \model{}.}

As shown in \cref{tab:module_ablation_app}, the performance of \model{} on MINERVA scales steadily with the strength of the backbone MLLM. Using the lightweight Qwen3-VL-8B yields 41.2\% accuracy ($2.0\%$ improvements compared to the direct inference), while swapping in stronger general-purpose models such as Gemini-2.5-Flash and OpenAI-o3 improves accuracy to 56.9\% and 59.0\%, respectively. 
The best results are obtained with Gemini-2.5-Pro (65.6\%), indicating that richer reasoning and instruction-following capabilities at the backbone level directly translate into better planning, evidence selection, and reflection for complex multi-hop queries.
At the same time, \model{} delivers consistent gains across a wide spectrum of MLLMs, suggesting that our \model{} framework is broadly applicable and can flexibly exploit future backbone improvements.

\begin{table}[t]
  \centering
  \resizebox{0.8\linewidth}{!}{
  \setlength{\tabcolsep}{6pt}
  \begin{tabular}{lc}
    \toprule
    \textbf{Backbone MLLM} & \textbf{MINERVA (Acc.\ \%)} \\
    \midrule
    Qwen3-VL-8B          & 41.2    \\
    Gemini-2.5-Flash     & 56.9  \\
    OpenAI-o3            & 59.0  \\
    Gemini-2.5-Pro       & \textbf{65.6} \\
    \bottomrule
  \end{tabular}
  }
  \vspace{-0.35em}
  \caption{\textbf{Backbone MLLM selection within \model{}.} The performance of \model{} on MINERVA scales steadily with the strength of the backbone MLLM. }
  \label{tab:module_ablation_app}
\end{table}

\paragraph{Structured vs. Unstructured Evidence List.}
As shown in \cref{tab:evidence_list_ablation}, replacing our structured, time-aligned evidence list with an unstructured flat list degrades performance on both benchmarks, indicating that temporally and semantically organized evidence is crucial for effective planning and reflection.

\begin{table}[t]
  \centering
  \resizebox{0.9\linewidth}{!}{
  \setlength{\tabcolsep}{5pt}
  \begin{tabular}{lcc}
    \toprule
    \textbf{Evidence Format} & \textbf{MINERVA} & \textbf{LVBench} \\
    \midrule
    Unstructured List  & 63.2 & 71.2 \\
    \textbf{Structured Evidence List (Ours)} & \textbf{65.6} & \textbf{74.8} \\
    \bottomrule
  \end{tabular}
  }
  \vspace{-0.4em}
  \caption{\textbf{Ablation on structured evidence list.}  
  Replacing our structured, time-aligned evidence list with an unstructured flat list hurts performance on both benchmarks, showing that organizing evidence by temporal and semantic grounding is important for effective planning and reflection.}
  \label{tab:evidence_list_ablation}
\end{table}

\paragraph{Confidence Threshold Sensitivity Analysis.}

As shown in \cref{tab:confidence_ablation}, a moderate confidence threshold yields the strongest results on MINERVA and ties for best performance on LVBench. Lower thresholds lead to premature halting and reduced accuracy, while overly strict thresholds offer no additional gains. This suggests that \model{} benefits from a balanced stopping criterion, confident enough to avoid early termination, yet flexible enough to prevent unnecessary observation rounds.

\begin{table}[t]
  \centering
  \resizebox{0.9\linewidth}{!}{
  \setlength{\tabcolsep}{8pt}
  \begin{tabular}{ccc}
    \toprule
    \textbf{Confidence Threshold} & \textbf{MINERVA} & \textbf{LVBench} \\
    \midrule
    0.5 & 64.2 & 73.2 \\
    0.7  & \textbf{65.6} & \textbf{74.8} \\
    0.9 & 65.4 &  \textbf{74.8} \\
    \bottomrule
  \end{tabular}
  }
  \vspace{-0.4em}
  \caption{\textbf{Ablation on confidence threshold.} 
  We vary the confidence threshold for halting, observing that different values trade off answer conservativeness and coverage on both benchmarks.}
  \label{tab:confidence_ablation}
\end{table}

\noindent\textbf{Extended Component Ablation.}
To verify that each component's contribution generalizes beyond the Gemini backbone, we repeat the component ablation (Tab.~\ref{tab:component_ablation}) using OpenAI-o3 on MINERVA.
Using only the Observer achieves 54.2\% accuracy; adding the Planner yields a 3.2\% improvement, and incorporating the Reflector provides a further 1.6\% gain.
These additive improvements are consistent with the Gemini-based results in Tab.~\ref{tab:component_ablation}, confirming that each \model{} component contributes meaningfully regardless of the backbone MLLM.

\subsection{Amortized Multi-Query Efficiency}
\label{sec:amortized}

One potential advantage of caption-based methods is that the video-level database can be reused across multiple queries for the same video, amortizing the construction cost. We evaluate this on LVBench under a multi-query setting (avg.\ 15 queries/video). DeepVideoDiscovery spends $637.2$\,s on database construction and an additional $2329.5$\,s on per-query localization and answering across all queries. 
By skipping the captioning stage entirely, \model{} answers all queries in $2179.5$\,s, achieving a \textbf{26.5\% reduction} in total inference time even in this setting favorable to database-reuse methods.
This result confirms that the efficiency gains of active, query-driven perception hold not only per-query but also in amortized multi-query scenarios.

\subsection{Token Efficiency Across Backbones}
\label{sec:token_efficiency}

\model{}'s token efficiency arises from how it adaptively plans for temporal and spatial token usage, rather than backbone-specific visual compression. 
Under the same OpenAI-o3 backbone, \model{} averages 62.5K input tokens per query on LVBench, only 5.8\% of DVD's usage and even lower than the Gemini-based \model{} (132.5K). This result demonstrates that \model{}'s efficiency does not rely on a specific backbone.

\subsection{Robustness Across Video Domains}
\label{sec:robustness}

To assess whether \model{}'s improvements hold across diverse video domains, we break down MINERVA results by domain category.
As shown in Tab.~\ref{tab:robustness_domain}, \model{} consistently outperforms Gemini-2.5-Pro across sports, instructional videos, and films, indicating that the gains from active perception are not specific to a particular video type.

\begin{table}[h]
\centering
\begin{tabular}{lccc}
\toprule
 & Sports & Instructional & Films \\
\midrule
Gemini-2.5-Pro & 57.5 & 57.2 & 75.2 \\
AVP (ours) & \textbf{60.8} & \textbf{59.6} & \textbf{77.1} \\
\bottomrule
\end{tabular}
\caption{Domain-level accuracy (\%) on MINERVA. \model{} improves over its Gemini-2.5-Pro backbone across all video domains.}
\label{tab:robustness_domain}
\end{table}

\section{Additional Qualitative Results}
\label{sec:appenddix_qualitative}

\begin{figure*}[t]
  \centering
  \includegraphics[width=0.9\linewidth]{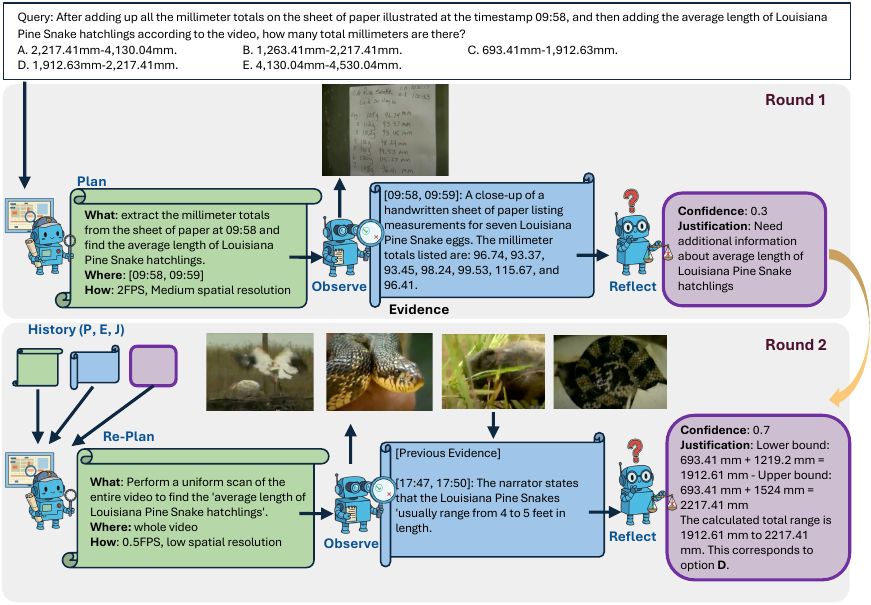}
\caption{\textbf{Qualitative example of multi-round active perception in \model{} (MINERVA sample).}
Given the query, ``After adding up all the millimeter totals on the sheet of paper illustrated at 09:58, and then adding the average length of Louisiana Pine Snake hatchlings according to the video, how many total millimeters are there?'', \model{} first plans to focus on the local timestamped frame at 09{:}58 and extracts the seven millimeter totals from the handwritten measurement sheet (Round~1). 
The reflector correctly judges that this evidence is insufficient because the average hatchling length is still unknown, and triggers a second round. 
In Round~2, the planner re-directs the observer to uniformly scan the full video at low FPS, locating a narrated segment that states hatchlings ``usually range from 4 to 5 feet in length.'' 
By fusing the previous numeric evidence with this newly discovered range, the reflector computes the total millimeter interval and selects the correct option.}
\label{fig:snake_case}
\end{figure*}

\subsection{Additional Visualization}

As illustrated in \cref{fig:snake_case}, this example showcases how \model{} leverages iterative planning to solve compositional, numerically precise queries that cannot be answered from a single view of the video.
In the first round, the agent executes a narrowly targeted observation around the specified timestamp to read off the millimeter totals from the paper, but the reflector explicitly flags that the evidence is incomplete.
The planner then revises its strategy, broadening the search space to a coarse scan over the entire video to hunt for the missing semantic attribute (the average hatchling length), which the observer recovers from narration.
Only after both local numeric measurements and global semantic context are available does the reflector combine them into the final answer.
This visualization shows \model{} could tackle complex, multi-hop video reasoning via its iterative design.

\subsection{Failure Case}

\begin{figure*}[t]
  \centering
  \includegraphics[width=0.8\linewidth]{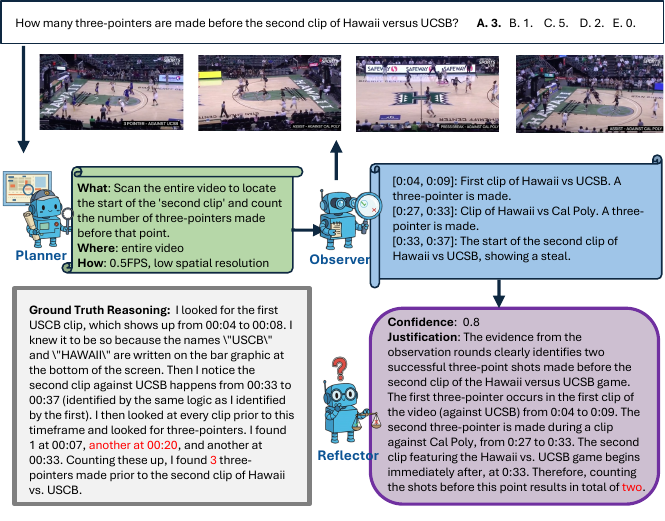}
\caption{\textbf{Failure Case of \model{}} (MINERVA sample). Given a long broadcast basketball video, \model{} must answer:
“How many three-pointers are made before the second clip of Hawaii versus UCSB?” 
The planner chooses to scan the entire video at \mbox{0.5 FPS} with low spatial resolution, the observer summarizes the retrieved segments into a structured evidence list, and the reflector produces a confident answer of two. 
However, the ground-truth reasoning (yellow box) shows that a three-pointer at 00{:}20 is missed, so the correct count is three. 
Although the internal reasoning over the collected evidence is coherent, the initial coarse observation policy fails to capture a short, local event, leading to an overconfident but incorrect prediction. }

\label{fig:failure}
\end{figure*}

In \cref{fig:failure}, we analyze a representative failure mode of \model{} on a fine-grained counting query.
To save computation, the planner opts for a coarse 0.5 FPS scan of the entire video and the observer only records two three-point plays before the second Hawaii–UCSB clip.
Since the missing shot at 00:20 is never observed, the reflector receives a logically consistent but incomplete evidence list and confidently outputs the wrong answer.
This case illustrates that, while our active perception pipeline is effective for locating dispersed, high-level evidence, it might make mistakes on questions that hinge on short, local events and subtle broadcast cues (e.g., bar graphics and rapid scoring plays).
We further analyze 50 randomly sampled LVBench failures, categorizing errors into \emph{planner}, \emph{observer}, and \emph{reflector}.
Only $8\%$ and $16\%$ of errors stem from planning and reflection, while most arise from observer perception errors ($76\%$).
This verifies the \model{}’s planner and reflector are relatively robust.

\section{Additional Implementation Detail}
\label{sec:appendix_implement}

\subsection{Prompts}

We provided the planner prompt, the observer prompt, and the reflector prompt as follow.  

\begin{tcolorbox}[promptstyle,breakable,title={Planner prompt (initial planning)}]
\footnotesize
\textbf{Function.}
\texttt{get\_planning\_prompt(query, video\_meta, options)}.\\
\textbf{Goal.}
You are an expert video analysis planner. Create a concise, observation plan to answer the user's query.

\vspace{0.35em}
\textbf{Inputs.}
\begin{itemize}[leftmargin=*]
  \item \textbf{User Query:} \texttt{\{full\_query\}}
  \item \textbf{Video Information:} duration in seconds
        (e.g., \texttt{Duration: \{duration\} seconds})
  \item \textbf{Options (optional):} multiple-choice options attached to the query
\end{itemize}

\vspace{0.35em}
\textbf{Planning framework.}
Produce observation with:
\begin{itemize}[leftmargin=*]
  \item \textbf{What (Reasoning Objective):} what the step tries to accomplish.
  \item \textbf{Where:} temporal span to examine, either
        \emph{uniform} (entire video) or a specific time range.
  \item \textbf{How:} \texttt{fps} and \texttt{spatial\_token\_rate}.
\end{itemize}

\vspace{0.35em}

\vspace{0.35em}
\textbf{Timestamp handling.}
First classify the query:
\begin{itemize}[leftmargin=*]
  \item \textbf{Factual questions:} e.g., “what”, “how many”, “who”, “which”,
        “count”, “identify”.
  \item \textbf{Reasoning / explanation questions:} e.g., “why”, “how”, “explain”,
        “reason”, “cause”.
\end{itemize}
Then apply:
\begin{itemize}[leftmargin=*]
  \item \textbf{Rule 1 (Exact ranges).}
        \begin{itemize}[leftmargin=*]
          \item Factual: use the \emph{exact} range, no padding
                (e.g., “07:15–07:18” $\rightarrow$ \texttt{[435.0, 438.0]}).
          \item Reasoning: add 15–30\,s padding before and after
                (e.g., “07:15–07:18” $\rightarrow$ \texttt{[420.0, 453.0]}).
        \end{itemize}
  \item \textbf{Rule 2 (Single timestamp).}
        \begin{itemize}[leftmargin=*]
          \item Factual: 1\,s forward window from timestamp
                (e.g., “at 02:15” $\rightarrow$ \texttt{[135.0, 136.0]}).
          \item Reasoning: add 15–30\,s context
                (e.g., “at 02:15” $\rightarrow$ \texttt{[120.0, 150.0]}).
        \end{itemize}
  \item \textbf{Rule 3 (Approximate / vague timing).}
        Use a $\pm$15\,s window around the mentioned time
        (e.g., “around 1:23” $\rightarrow$ \texttt{[68.0, 98.0]}).
\end{itemize}

\vspace{0.35em}
\textbf{Heuristics for unknown timing.}
\begin{itemize}[leftmargin=*]
  \item “opening / beginning” $\rightarrow$ \texttt{[0, 30]}.
  \item “end / ending” $\rightarrow$
        \texttt{[max(0, duration - 30), duration]}.
  \item No timing mentioned: use a coarse uniform scan
        with \texttt{fps} in 0.25–1.0 and low/medium resolution.
\end{itemize}

\vspace{0.35em}
\textbf{Step configuration guidelines.}
\begin{itemize}[leftmargin=*]
  \item \textbf{Uniform scan (timing unknown).}
        \texttt{load\_mode = "uniform"}, \texttt{fps} in 0.25–1.0,
        \texttt{spatial\_token\_rate} $\in$ \{\texttt{"low"}, \texttt{"medium"}\},
        \texttt{regions = []}.
  \item \textbf{Region analysis (explicit timestamps).}
        \texttt{load\_mode = "region"}, \texttt{fps} $\approx$ 2.0,
        \texttt{spatial\_token\_rate} $\in$ \{\texttt{"low"}, \texttt{"medium"}\},
        \texttt{regions = [[start, end]]}.
\end{itemize}

\vspace{0.35em}
\textbf{Few-shot examples.} [Few\_examples]

\vspace{0.35em}
\textbf{Output format.}
Return a single JSON object with:
\begin{itemize}[leftmargin=*]
  \item \texttt{reasoning}: natural language explanation of your planning.
  \item \texttt{plans}: 
        \texttt{what} $\in$ \texttt{sub\_query},
        \texttt{where} $\in$ \{\texttt{"uniform"}, \texttt{"region"}\},
        \texttt{how} $\in$ numeric \texttt{fps} (0.1–5.0), \texttt{spatial\_token\_rate}
        $\in$ \{\texttt{"low"}, \texttt{"medium"}\}, and \texttt{regions}
        (list of \texttt{[start, end]} in seconds; empty for uniform).
\end{itemize}
\end{tcolorbox}

\begin{tcolorbox}[promptstyle,breakable,title={Observer prompt (video inference / evidence extraction)}]
\footnotesize

\textbf{Goal.}
Analyze a specific video segment and extract precise, time-stamped evidence relevant to the user query.

\vspace{0.4em}
\textbf{Inputs.}
\begin{itemize}[leftmargin=*]
  \item \textbf{sub\_query:} the focused question for this round.
  \item \textbf{original\_query:} the full user question (for multi-step agents).
  \item \textbf{context:} accumulated evidence from previous rounds.
  \item \textbf{start\_sec, end\_sec:} bounds of the segment to analyze.
  \item \textbf{video\_duration\_sec:} duration of the full video.
  \item \textbf{is\_region:} whether this step analyzes a specified region or uniform scan.
  \item \textbf{regions:} list of \texttt{[start, end]} spans if multiple clips are provided.
\end{itemize}

\vspace{0.4em}
\textbf{Prompt structure.}
\begin{itemize}[leftmargin=*]
  \item Primary task: describe visually relevant events in the analyzed video span.
  \item Provide:
    \begin{itemize}[leftmargin=*]
      \item \textbf{Detailed observations} tied to the query.
      \item \textbf{Key timestamp ranges} (\texttt{timestamp\_start}, \texttt{timestamp\_end}) for each salient event.
      \item \textbf{Reasoning} connecting observations to the sub-query.
    \end{itemize}
\end{itemize}

\vspace{0.4em}
\textbf{Timestamp and evidence rules.}
\begin{itemize}[leftmargin=*]
  \item Round timestamps to \textbf{integer seconds}: floor(start), ceil(end).
  \item List \emph{all} relevant intervals for events that may match the query.
  \item Use context to avoid redundant descriptions.
\end{itemize}

\vspace{0.4em}
\textbf{Multiple-clip handling.}
\begin{itemize}[leftmargin=*]
  \item When inputs include several regions, each corresponds to its absolute
        time span in the original video.
  \item You may reference clips descriptively (e.g., “Clip 1”, “Clip 2”).
\end{itemize}

\vspace{0.4em}
\textbf{Fallback rule (critical).}
If analyzing a \emph{region} and no relevant information is present:
\begin{itemize}[leftmargin=*]
  \item Explicitly state: “No relevant information found in this time segment.”
  \item Suggest expanding search to a uniform scan or additional regions.
\end{itemize}

\vspace{0.4em}
\textbf{Output format.}
Return a JSON object:
\begin{verbatim}
{
  "detailed_response": "...",
  "key_evidence": [
    {
      "timestamp_start": <number>,
      "timestamp_end": <number>,
      "description": "..."
    }
  ],
  "reasoning": "..."
}
\end{verbatim}

\vspace{0.3em}
\textbf{Example.} [Few\_examples]
\end{tcolorbox}

\begin{tcolorbox}[promptstyle,breakable,
  title={Reflector prompt (evidence sufficiency checker)}]
\footnotesize

\textbf{Goal.}
Given the original query and cumulative evidence from all observation rounds,
decide whether the current evidence is sufficient to answer the query, and
produce a justification that either (i) contains the final answer, or
(ii) explains what is missing.

\vspace{0.35em}
\textbf{Inputs.}
\begin{itemize}[leftmargin=*,itemsep=2pt,topsep=2pt]
  \item \textbf{query:} original user query (with options if MCQ).
  \item \textbf{evidence\_summary:} aggregated evidence from all Observer steps.
  \item \textbf{video\_duration:} total duration in seconds.
  \item \textbf{options:} optional list of MCQ options.
\end{itemize}

\textbf{Your task.}
\begin{itemize}[leftmargin=*,itemsep=2pt,topsep=2pt]
  \item Decide a boolean \texttt{sufficient} indicating whether the evidence
        is enough to answer the query.
  \item Output a calibrated \texttt{confidence} $C \in [0,1]$ for the
        sufficiency decision. This is the score compared against
        $\tau_{\text{conf}}$ in \cref{alg:avp_min_history}: as a guideline, $C \ge 0.7$ when
        the evidence clearly entails an answer, $0.4 \le C < 0.7$ when it is
        suggestive but incomplete, and $C < 0.4$ when the key cues are
        unaddressed.
  \item \textbf{If sufficient (true):} the justification must give the
        \emph{direct answer}.
        \begin{itemize}[leftmargin=*,itemsep=1pt,topsep=1pt]
          \item MCQ: state the option letter (A/B/C/...) and a brief reason.
          \item Open-ended: clearly state the answer in natural language.
        \end{itemize}
  \item \textbf{If not sufficient (false):} the justification must explain
        what information is missing or uncertain (e.g., which regions,
        entities, or temporal spans require additional observation).
  \item Always provide a short \texttt{reasoning} paragraph that summarizes
        why the evidence is (not) sufficient.
\end{itemize}

\vspace{0.35em}
\textbf{Required JSON output (LLM response).}
\begin{verbatim}
{
  "sufficient": <true | false>,
  "confidence": <float in [0, 1]>,
  "justification": "...",
  "reasoning": "..."
}
\end{verbatim}

\textbf{Few-shot examples.} [Few\_examples]

\end{tcolorbox}